%% file: main.tex
\documentclass[runningheads]{llncs}
\usepackage{eccv}
\usepackage{eccvabbrv}
\usepackage{graphicx}
\usepackage{booktabs}
\usepackage{algorithm}
\usepackage{algorithmicx}
\usepackage{algpseudocode}
\usepackage[utf8]{inputenc}
\usepackage[T1]{fontenc}
\usepackage{amsfonts}
\usepackage{nicefrac}
\usepackage{microtype}
\usepackage{xcolor}
\usepackage{comment}
\usepackage{arydshln}
\usepackage{enumitem}
\usepackage[normalem]{ulem}
\usepackage{colortbl}
\usepackage{nicematrix}
\usepackage{lipsum}
\usepackage{cancel}
\usepackage{tabularx}
\usepackage{multicol}
\usepackage{multirow}
\usepackage{amssymb}
\usepackage{bm}
\usepackage{subcaption}
\usepackage{wrapfig}
\usepackage{indentfirst}
\usepackage[accsupp]{axessibility}
\usepackage{hyperref}
\usepackage{orcidlink}

\makeatletter
\newcommand{\printfnsymbol}[1]{%
  \textsuperscript{\@fnsymbol{#1}}%
}
\renewcommand*{\@fnsymbol}[1]{\ifcase#1\or*\or$\dagger$\or\else\@arabic{#1}\fi}
\makeatother

\begin{document}

\title{Online Versatile Incremental Learning:\\Towards Class and Domain-Agnostic Adaptation at Any Time}

\titlerunning{Online Versatile Incremental Learning}

\author{
Jae-Ho Lee\inst{1}\orcidlink{0009-0003-0960-824X} \and
Min-Yeong Park\inst{2}\thanks{Work done at Korea University.}\orcidlink{0009-0007-4143-7283} \and
Jun-Yeong Moon\inst{1}\orcidlink{0000-0002-0543-2588} \and \\
Jung Uk Kim\inst{2}\thanks{Corresponding Author.}\orcidlink{0000-0003-4533-4875} \and
Gyeong-Moon Park\inst{1}\printfnsymbol{2}\orcidlink{0000-0003-4011-9981}
}

\authorrunning{J.-H. Lee et al.}

\institute{Korea University, Seoul, Republic of Korea \\
\email{\{jaeho-lee, moonjunyyy, gm-park\}@korea.ac.kr} \and
Kyung Hee University, Yongin, Republic of Korea \\
\email{\{pmy0792, ju.kim\}@khu.ac.kr}}

\maketitle
\input{sec/0_abstract}
\input{sec/1_intro}
\input{sec/2_related_work}
\input{sec/3_method}

\input{sec/4_experiment}
\input{sec/5_conclusion}
\clearpage
\section*{Acknowledgements}
This work was partly supported by the Institute of Information \& Communications Technology Planning \& Evaluation (IITP) grant funded by the Korea government (MSIT) (No.RS-2026-25507543, Development of AI Co-Scientist based on Scientific Causal World Model, 80\%), and in part by Korea Planning \& Evaluation Institute of Industrial Technology (KEIT) grant funded by the Korea government (MOTIE) (RS-2024-00444344), and by the “Advanced GPU Utilization Support Program” funded by the Government of the Republic of Korea (Ministry of Science and ICT).
\bibliographystyle{splncs04}
\bibliography{main}
\end{document}

%% file: sec/0_abstract.tex
\begin{abstract}
Continual learning enables vision systems to adapt to ever-changing data distributions. Despite significant advances, existing approaches fail to capture continuous and concurrent shifts in classes and domains, a critical capability for real-world deployment. This work introduces \textbf{Online VIL (Online Versatile Incremental Learning)}, a novel scenario where class concepts and visual domains evolve simultaneously online without explicit boundaries. To better adapt to the challenges of such dynamic environments that more closely resemble real-world conditions, we propose a novel framework \textbf{TopFlow}, \textbf{Top}ology preservation with \textbf{Flow} matching representation that contains two complementary mechanisms: \textbf{Domain-agnostic Flow Matching (DFM)} and \textbf{Global Topology Preservation (GTP)}. DFM guides the model to have domain-agnostic representations by integrating the geodesic flow kernel into contrastive learning. In contrast, GTP maintains the global structure of the feature space without explicitly storing past examples. Our extensive experiments demonstrate that TopFlow effectively addresses the limitations of existing methods within the Online VIL scenario, achieving state-of-the-art performance in challenging Online VIL. The proposed methods suggest potential directions for building continual learning systems in realistic dynamic environments. Our implementation code is available at \url{https://github.com/KU-VGI/Online-VIL}.
\keywords{Online learning \and Incremental learning \and Real-world scenario}
\end{abstract}

%% file: sec/1_intro.tex
\section{Introduction}
\label{sec:intro}

\begin{figure}[!t]
\centering
\begin{tabular}{@{}c@{}}
    \begin{subfigure}[t]{.245\columnwidth}
        \includegraphics[width=1.0\columnwidth]{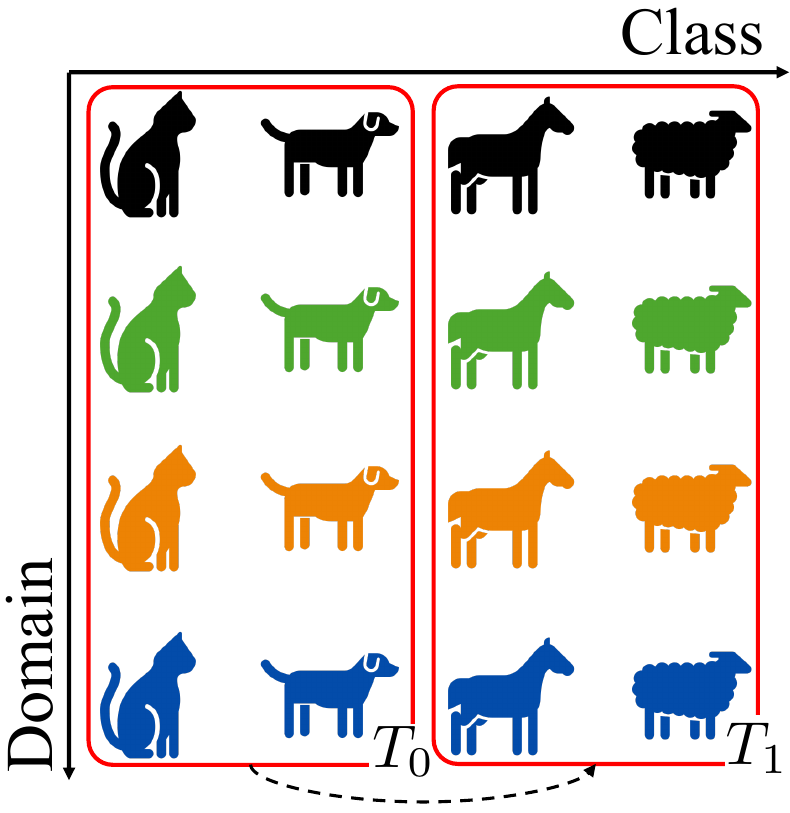}
        \caption{CIL.}
        \label{fig:cil}
    \end{subfigure}
    \begin{subfigure}[t]{.245\columnwidth}
        \includegraphics[width=1.0\columnwidth]{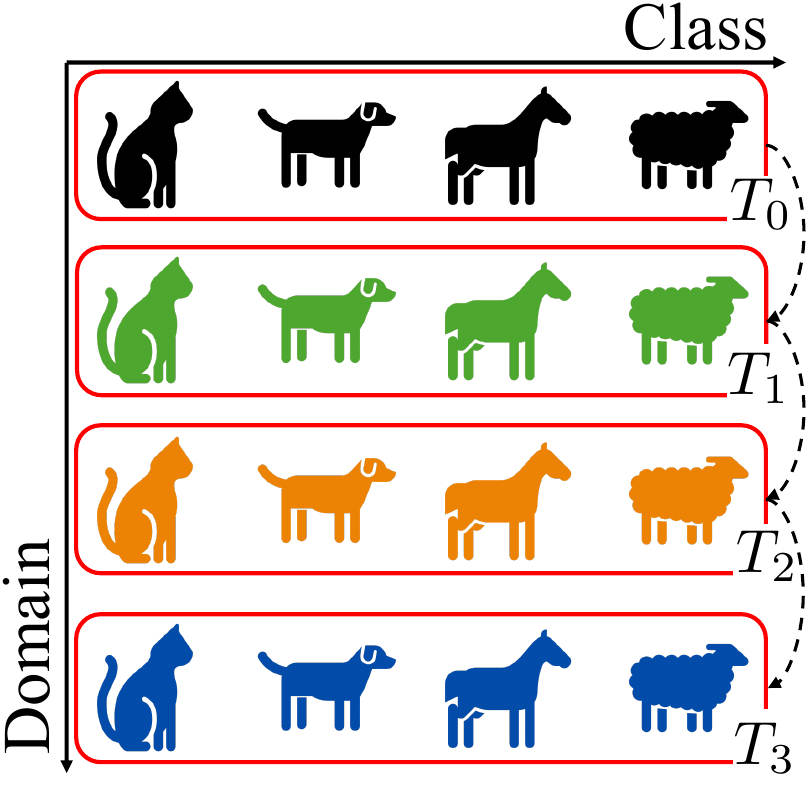}
        \caption{DIL.}
        \label{fig:dil}
    \end{subfigure}
    \begin{subfigure}[t]{.245\columnwidth}
        \includegraphics[width=1.0\columnwidth]{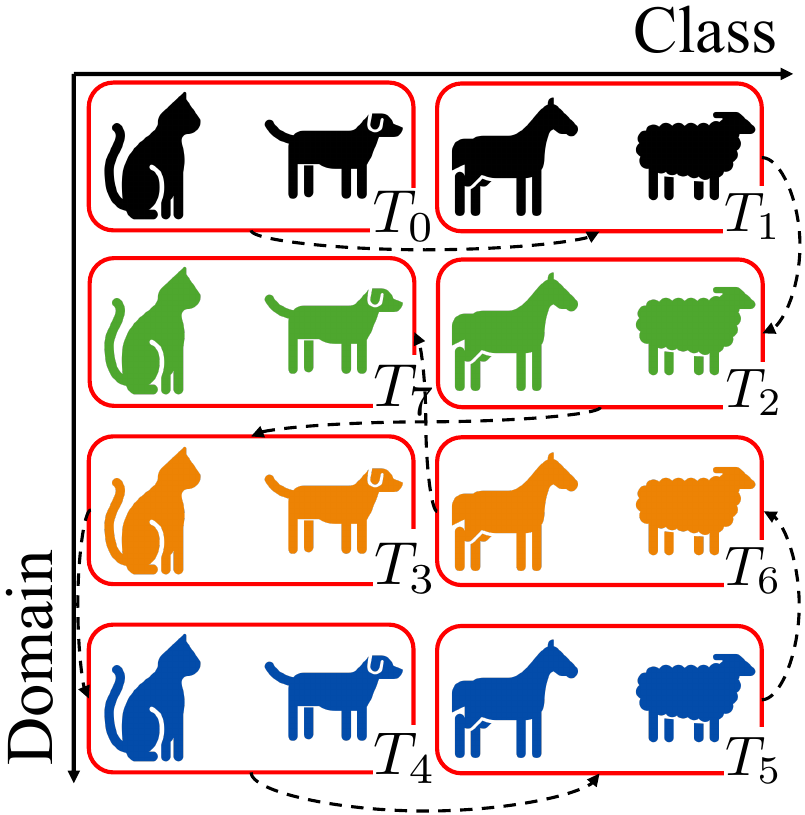}
        \caption{VIL.}
        \label{fig:vil}
    \end{subfigure}
    \begin{subfigure}[t]{.25\columnwidth}
        \includegraphics[width=1.0\columnwidth]{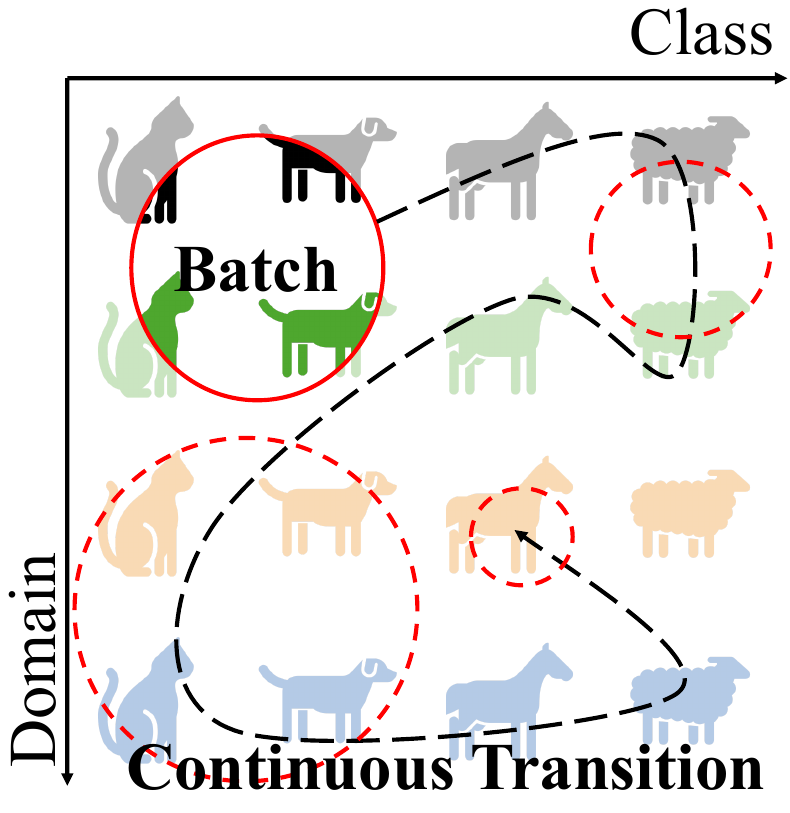}
        \caption{{\textbf{Online VIL (Ours).}}}
        \label{fig:onlinevil}
    \end{subfigure}
\end{tabular}
\caption{Conceptual comparison of (a) Class Incremental Learning (CIL), (b) Domain Incremental Learning (DIL), (c) Versatile Incremental Learning (VIL), and (d) Online Versatile Incremental Learning (Online VIL, Ours). Red lines indicate the scopes observable at once, while black dashed lines depict transitions across time.}
\label{fig:scenario}
\end{figure}

\noindent Continual Learning (CL) \cite{li_learning_2017,kirkpatrick2017overcoming,wang2022sprompts,wang2022learning,wang2022dualprompt,smith2023coda,zhang2023slca,park2024pre} has gained increasing attention as deep learning moves closer to the real world, where data distributions evolve. A key challenge is catastrophic forgetting, where adapting to new knowledge disrupts previous knowledge. To mitigate this, prior research has introduced distinct paradigms such as Class Incremental Learning (CIL) and Domain Incremental Learning (DIL), targeting a specific form of non-stationarity. More recently, Online Continual Learning (OCL) \cite{wang2023cba,wei2023online,he2024dyson} has been proposed to handle streaming data under memory and single-pass constraints, including scenarios with blurry task boundaries \cite{koh2021online,bang2021rainbow,moon2023online}. However, most works still rely on CIL or DIL, limiting their ability to capture the full complexity of real-world dynamics.
\par\smallskip
\noindent A recently introduced Versatile Incremental Learning (VIL) \cite{park2024versatile} suggests a more realistic scenario where new tasks have a broader chance to evolve in both directions of the classes and domains, without prior knowledge. While VIL marks a step toward realism, it assumes discrete increments that provide implicit structural information about distribution shifts. In contrast, real-world environments exhibit continuous transitions without boundaries, offering no such organizational cues. Moreover, environmental change in reality is continuous and unpredictable, due to locally constrained information and multi-factor interactions. For instance, an autonomous driving system may suddenly face both a new object and a shift in weather or city conditions, requiring immediate (online) adaptation without knowing whether it concerns classes, domains, or both.
\par\smallskip
\noindent To this end, we introduce \textbf{Online} \textbf{V}ersatile \textbf{I}ncremental \textbf{L}earning (\textbf{Online VIL}), a new scenario that captures both the unpredictable heterogeneous shifts in an online manner. Online VIL is distinguished by reflecting the evolution of the natural information stream, characterized by unpredictable, gradual, and heterogeneous shifts that occur along multiple evolutionary trajectories. As shown in Figure \ref{fig:scenario}, Online VIL allows flexible transitions across class and domain dimensions; each sequence presents distinct challenges without heuristic patterns. Consequently, Online VIL enables faithful evaluation of CL models and provides a foundation for systems that operate under real-world dynamics.
\par\smallskip
\noindent In the Online VIL scenario, adaptation to chaotic shifts in classes and domains without explicit access to prior inputs is crucial for distinguishing class-discriminative knowledge from domain-specific knowledge. Otherwise, models rely on the spurious features of current distributions and tend to exhibit rapid forgetting and a lack of generalization. Through systematic layer-wise feature analysis of pre-trained Vision Transformers, we observe that early layers predominantly capture domain-specific patterns (e.g., texture, lighting), while deeper layers encode class-discriminative features and more abstract semantic representations. This motivates a novel \textbf{Domain-agnostic Flow Matching (DFM)} technique, which aligns features with the intrinsic geometry of pre-trained knowledge through reconceptualized geodesic flow kernel \cite{gong2012geodesic} while mitigating domain-specific shifts.
\par\smallskip
\noindent In addition to class and domain-agnostic alignment, preserving the global topology of learned features is essential for maintaining semantic continuity across evolving tasks. Conventional objectives often fail to maintain the structural relationships of feature space because they shrink the occupation of missing classes in the feature space. To address this, we propose \textbf{Global Topology Preservation (GTP)}, which preserves invariant geometric configurations in feature space, enabling robust knowledge retention without requiring complete class coverage.
\par\smallskip
\noindent Integrating these components, we introduce \textbf{Top}ology preservation with \textbf{Flow} matching representation (\textbf{TopFlow}), a novel framework designed for Online VIL. With recognition of DFM and GTP regularization for geometry, TopFlow ensures robust adaptation to unpredictable shifts. To evaluate its effectiveness, we conduct extensive experiments in Online VIL and observe that TopFlow consistently outperforms existing state-of-the-art methods across multiple benchmarks. Our contributions are summarized as follows:
\begin{itemize}
    \item We introduce \textbf{Online VIL}, a realistic scenario for evaluating continual learning where class and domain distributions evolve continuously and jointly with ambiguous task boundaries.
    \item We reveal a novel role of the pre-trained ViT layer that encodes class and domain knowledge. Leveraging this insight, we propose Domain-agnostic Flow Matching (\textbf{DFM}) to learn domain-agnostic representations by integrating the geodesic flow kernel into contrastive learning.
    \item We propose Global Topology Preservation (\textbf{GTP}), a mechanism for maintaining knowledge representations using feature topologies without explicit memory of previous inputs.
    \item We demonstrate that the proposed \textbf{TopFlow} significantly outperforms existing state-of-the-art methods through comprehensive experiments in our challenging Online VIL scenario.
\end{itemize}

%% file: sec/2_related_work.tex
\section{Related Work}
\label{sec:relatedwork}
\noindent \textbf{Online Continual Learning.} Online Continual Learning (OCL) \cite{de2021continual,gunasekara2023survey} has emerged as a pragmatic paradigm that reflects the challenges of real-world settings, where data arrive as continuous streams and models must operate under minimal batch sizes, single-pass constraints, and strict computational and memory limitations. Traditionally, OCL methods rely on replay buffers \cite{rolnick2019experience, mai2021supervised} to store a small subset of past data, thereby mitigating catastrophic forgetting while learning new tasks. Recent advances explore leveraging prototypes \cite{wei2023online}, replay-free strategies \cite{zajkac2024prediction}, and pre-trained models with prompt \cite{moon2023online}. While effective in class or domain increments, this dependence on memory limits scalability and realism. Methods addressing more realistic scenarios with blurry or ambiguous task boundaries have emerged \cite{koh2021online,bang2021rainbow}. However, the replay methods require growing memory in proportion to task diversity. Also, blurry boundary methods still assume either class-only or domain-only shifts, failing to capture the heterogeneous evolution of a realistic data stream. Therefore, we propose Online Versatile Incremental Learning (Online VIL). This scenario exposes models to unpredictable shifts in both classes and domains while enforcing online constraints that limit memory and multi-pass access.
\\
\\
\noindent \textbf{Geodesic Flow Kernel.} The Geodesic Flow Kernel (GFK) \cite{gong2012geodesic,gopalan2011domain} has been widely used in unsupervised domain adaptation to align feature distributions between a predefined source and target domain. Conventional applications approximate the geodesic on the Grassmannian manifold between two static domains, which requires that the data of the source and target domains are fully available and static. In contrast, Online VIL presents unique challenges: domain shifts occur continuously and unpredictably, and data arrive sequentially, making it impossible to estimate the flow offline. To address these challenges, we reformulate the approach from GFK and design a novel Domain-Agnostic Flow Matching (DFM). Unlike traditional geometry estimation in feature space, DFM is designed for sequentially arriving data and evolving domains without relying on holistic data access. Through both empirical and theoretical analysis of feature geometry, DFM enables online alignment on feature geometry in dynamic conditions, avoiding computationally expensive higher-order manifold computations. This design fundamentally extends the applicability and purpose of GFK, enabling robust domain-agnostic feature matching in the OCL.
\\
\\
\noindent \textbf{Feature Topology.} Several works have leveraged the topology of feature space to mitigate catastrophic forgetting in sequential learning. \cite{TOPIC} employs elastic Hebbian graphs to preserve neighborhood relationships during incremental updates, while \cite{BOCL} uses self-organizing maps to identify representative feature points and restrict their displacement. More recent approaches, such as \cite{liu2022model} and \cite{SOUL}, maintain pair-wise instance similarity or local topological relations by decomposing the global structure, further reducing forgetting across tasks. However, these methods are limited to offline CL, requiring full access to past samples to construct the feature topology. Furthermore, existing topology preservation methods rely on complete semantic information across all classes. In online scenarios where only partial class information is available at each time step, these methods suffer from incomplete topology construction, resulting in suboptimal feature space organization. In contrast, our work proposes a Global Topology Preservation (GTP) strategy that maintains structural knowledge without requiring the storage of previous data.

%% file: sec/3_method.tex
\section{Method}
\label{sec:method}
\setlength{\belowdisplayskip}{1pt} \setlength{\belowdisplayshortskip}{1pt}
\setlength{\abovedisplayskip}{1pt} \setlength{\abovedisplayshortskip}{1pt}

\subsection{Problem Setup: Online Versatile Incremental Learning}
\label{subsec:3.1}
\noindent In real-world scenarios, data distributions gradually evolve, often exhibiting significant variability across multiple dimensions such as spatial domains, semantic classes, and temporal shifts.
We propose a novel scenario termed Online Versatile Incremental Learning (Online VIL), which simulates these properties by constructing a continuous data stream with stochastically varying distributions. Given a dataset with $n_\mathcal{D}$domains and $n_\mathcal{C}$ classes, we define the product category space as follows:
\begin{align}
    \mathcal{X} &= \bigcup_{i=1}^{n_\mathcal{D}} \mathcal{D}_i,\;
    = \bigcup_{j=1}^{n_\mathcal{C}} \mathcal{C}_j,\;
    = \bigcup_{i=1}^{n_\mathcal{D}} \bigcup_{j=1}^{n_\mathcal{C}} \mathcal{K}_{i,j},\quad \text{where}\quad
    \mathcal{K}_{i,j} = \mathcal{D}_i \cap \mathcal{C}_j,
\end{align}
where $\mathcal{D}_i$ is a set of samples in a domain $i$ and $\mathcal{C}_j$ is in a class $j$, and $\mathcal{K}_{i,j}$ is samples that belong to domain $i$ and class $j$.
Machine learning typically assumes the data $\mathcal{X}$ as a lower-dimensional manifold embedded in a high-dimensional space \cite{cayton2005algorithms}.
In this context, we can conceptualize the data stream as a trajectory through the joint domain-class space, where each timestep $t$ provides an observation window $\mathcal{V}_t \subset \mathcal{X}$ which is a disjoint open neighborhood of data $\mathcal{X}$.
This geometric interpretation naturally leads to our manifold-based approach in the subsequent technical development.

Inspired by Si-Blurry \cite{moon2023online}, we divide categories into disjoint sets ($\mathcal{K}_\text{disjoint}$) with clear distributional boundaries and blurry sets ($\mathcal{K}_\text{blurry}$) with deliberately abstracted distributions.

Algorithm A.1 in Appendix details our task construction process, which proceeds in six main steps: (\textit{i}) \textit{category partitioning} to establish variable distributional clarity, (\textit{ii}) \textit{sample extraction} to create diverse distributional patterns, (\textit{iii}) \textit{sample redistribution} to simulate realistic category overlap, (\textit{iv}) \textit{randomized task assignment} to eliminate artificial task boundaries, (\textit{v}) \textit{task construction} with varying characteristics, and (\textit{vi}) \textit{batch generation} with constrained visibility windows.
The Online VIL scenario distinguishes itself from traditional CL scenarios through three key characteristics:
\begin{enumerate}[nosep, leftmargin=12pt]
\item \textbf{Locally Limited Visibility.}
    At any time step, the model observes only a small fraction in both number of samples ($|\mathcal{V}_t| \ll |\mathcal{D}|$) and categories ($\exists \mathcal{K}_{i,j}: x \in \mathcal{K}_{i,j} \land x \notin \mathcal{V}_t$), reflecting real-world constraints on data accessibility, combines both spatial and temporal restrictions.
\item \textbf{Continuous Smooth Variation.}
    The variation is smooth and continuous over time, making it hard to distinguish the distribution shift without a global context.
\item \textbf{Dynamic Distribution.}
    Each category appears and disappears gradually. The variance across spatial (domain appearances), semantic (class properties), and temporal (distribution evolution) dimensions requires simultaneous adaptation to new patterns and retention of previously acquired knowledge.
\end{enumerate}
This formulation demands adaptation mechanisms that extract meaningful patterns despite high variance and constrained observability.
Traditional CL methods, which assume either domain stability or class stability, fail to resemble these fundamental challenges in real-world perception systems.

\subsection{Domain-agnostic Flow Matching}
\label{sec:3.2}

\begin{figure}[!t]
\centering
\captionsetup[subfigure]{font=scriptsize,labelfont=scriptsize}
\begin{tabular}{cc}
    \begin{minipage}{.65\linewidth}
        \begin{subfigure}[t]{\linewidth}
            \includegraphics[width=\linewidth]{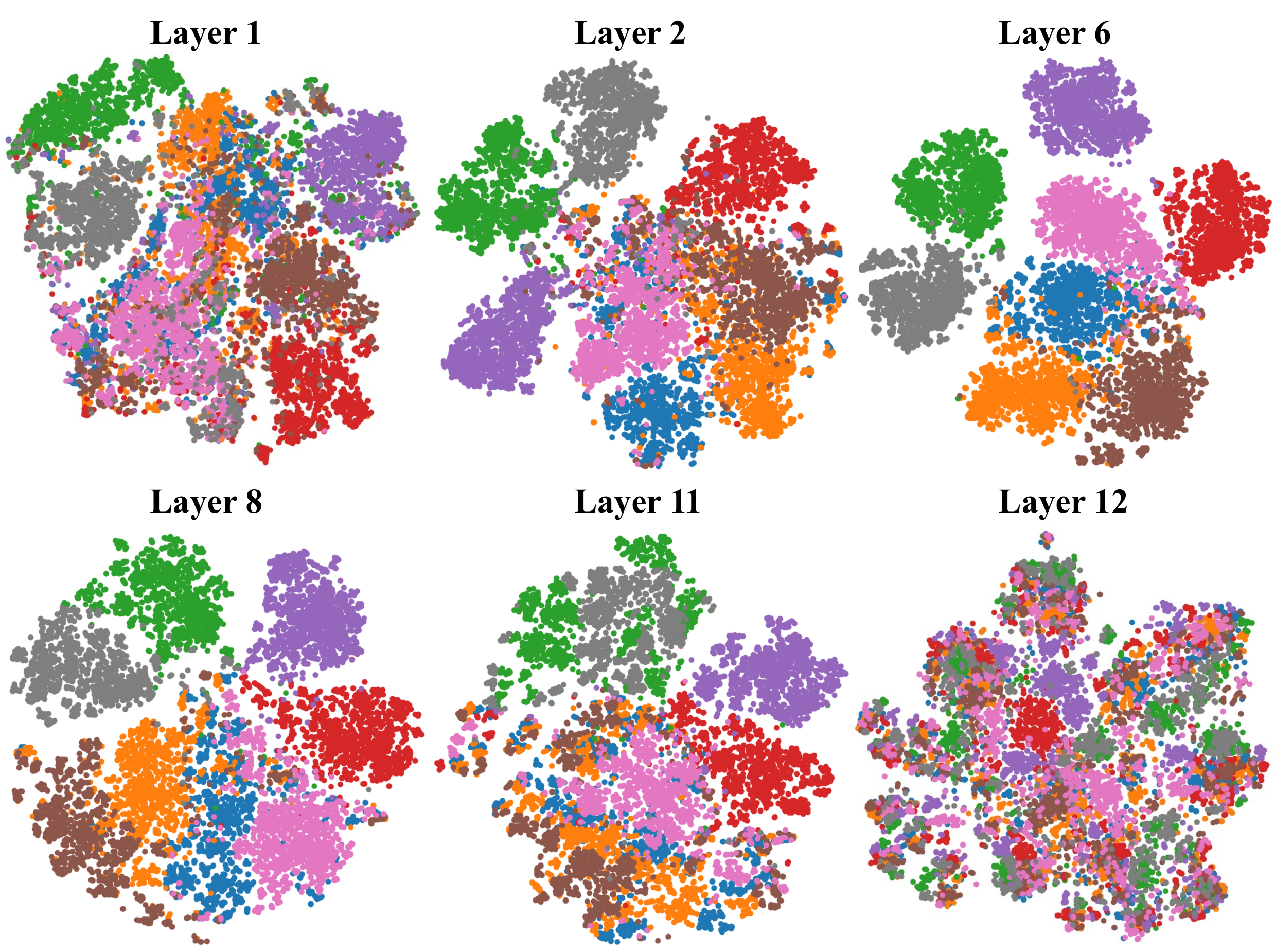}
            \caption{t-SNE visualization of features from pre-trained ViT.}
            \label{fig:tsne-a}
        \end{subfigure}
    \end{minipage}
    &
    \begin{minipage}{.37\linewidth}
        \begin{subfigure}[t]{\linewidth}
        \centering
            \includegraphics[width=.75\linewidth]{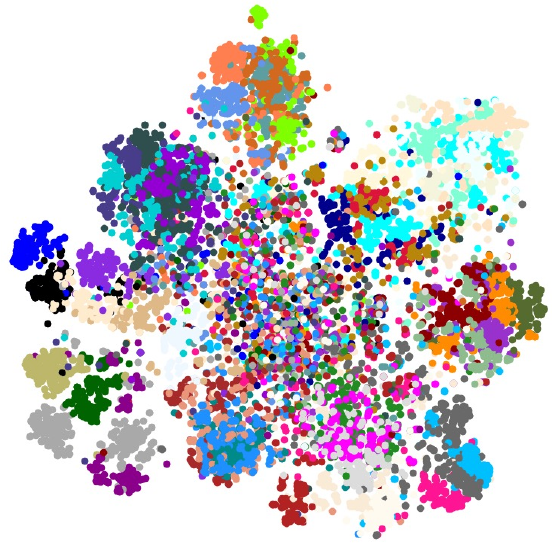}
            \caption{Class-wise coloring on last layer.}
            \label{fig:tsne-b}
        \end{subfigure}
        \begin{subfigure}[t]{\linewidth}
        \centering
            \includegraphics[width=.85\linewidth]{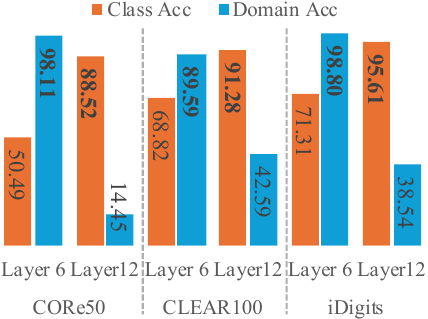}
            \caption{Linear Probing results.}
            \label{fig:acc_fig}
        \end{subfigure}
    \end{minipage}
\end{tabular}
\caption{Analysis of layer-wise knowledge on pre-trained ViT for data with changing distributions (CORe50). (a) t-SNE visualization of features from certain layers of the pre-trained ViT. The color indicates the domain. (b) The t-SNE visualization of the last layer with class-wise color. (c) Accuracy of linear probing for class and domain classification from different layers.}
\label{fig:tsne}
\end{figure}

\noindent The Online VIL scenario, with limited visibility, multi-dimensional variance, and dynamic distribution, poses challenges beyond traditional CL.
While the original VIL work measured feature similarity, it did not explicitly examine \emph{how} domain and class signals are encoded across the backbone under noisy and concurrent class/domain shifts.
This matters because the one-pass constraint and the lack of rehearsal make representations highly sensitive to the local batch composition, and thus prone to overfitting transient domain cues. To better understand this, we analyze the representations of \emph{frozen} pre-trained Vision Transformers, which have become the standard backbone in recent CL research \cite{wang2022sprompts,wang2022dualprompt,wang2022learning,smith2023coda,gao2023unified}.
Although not trained under Online VIL, a frozen backbone provides a stable proxy, reflecting common practice in CL of freezing the ViT and updating only a small number of parameters (e.g., prompts).
Its hierarchical representations largely dictate how domain-specific variations and class-level semantics interact.
Motivated by this, we conduct a layer-wise analysis, expecting deeper layers to align more closely with class semantics, as their outputs directly drive the classification head.

\begin{figure}[!t]
    \centering
    \includegraphics[width=0.95\linewidth]{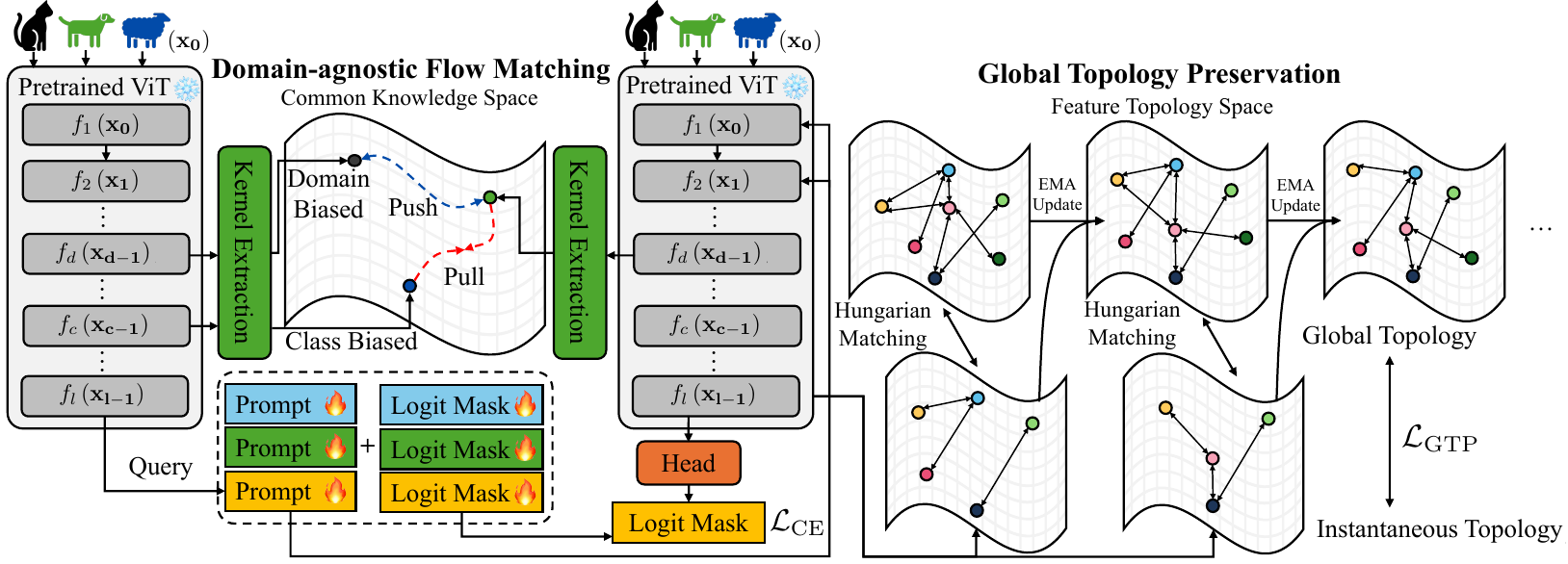}
    \caption{Architecture overview. TopFlow comprises two components: Domain-agnostic Flow Matching (DFM) and Global Topology Preservation (GTP). DFM promotes domain-agnostic representation accumulation, while GTP preserves the global feature topology.}
    \label{fig:mainl}
\end{figure}

\noindent We perform a t-SNE study and linear probing to investigate this.
Our t-SNE visualizations (Figures \ref{fig:tsne-a}, \ref{fig:tsne-b}) show that intermediate features group by \emph{domain}, while final features group by \emph{class}. Linear probing (Figure \ref{fig:acc_fig}) further confirms stronger domain discrimination in intermediate layers and stronger class discrimination in the final layer.
These findings reveal a \emph{structural representation gap} in Online VIL, motivating our proposed \textbf{Domain-agnostic Flow Matching (DFM)}: a geometry-informed contrastive loss that pulls the \emph{adaptable intermediate} representation toward the \emph{semantic} geometry of the final layer while pushing it away from domain-biased directions.
\par\smallskip
\noindent\textbf{From Geodesic-Flow Intuition to Layer-Wise Comparison Geometry.}
The Geodesic Flow Kernel (GFK) \cite{gong2012geodesic} suggests that features from a distribution can be represented by a subspace on the Grassmann manifold, and similarity can be computed through a comparison geometry defined over such subspaces.
In Online VIL, however, explicitly estimating \emph{inter-domain} subspaces is often ill-posed due to continuous, mixed, and unlabeled domain evolution.
Instead, we leverage the empirical layer-wise gap in Fig.~\ref{fig:tsne} and treat \emph{layers} as the unit of geometry: intermediate features are typically more domain-sensitive, while final features are more class-semantic.
Accordingly, we extract a \emph{batch-wise common comparison geometry} between intermediate and final representations, and use it to normalize similarities in a contrastive objective.
\noindent\textit{Residual-layer dynamics and local approximation.}
For convenience, we consider architectures with matched feature dimensionality across layers, such as ViTs \cite{dosovitskiy2020image}. With a cascade of functions with residual connections:
\begin{align}
    \bm{h}_0 = \bm{x},\quad
    \bm{h}_n = f_n(\bm{h}_{n-1}) + \bm{h}_{n-1},\quad
    n = 1, 2, \ldots, l,
\end{align}
where $f_i : \mathbb{R}^{b \times d} \rightarrow \mathbb{R}^{b \times d}$ is a function of the $i$-th layer, and $\bm{h}_l \in \mathbb{R}^{b \times d}$ is the last layer feature with batch size $b$ and feature dimension $d$.
With the local neighborhood $\mathcal{V}_t$ on a Riemannian manifold as mentioned in Section \ref{subsec:3.1}, we can take a first-order approximation with the infinitesimal variation of feature $\bm{h}_l$, as detailed in Equation A.3 and A.5.
Then, the inner product between $h_n$ and $h_l$ can be written as
\begin{align}
    \langle \bm{h}_n, \bm{h}_l \rangle
    &= \int_{X} \bm{h}_n^T \bm{h}_l d\bm{x} = \mathbb{E}_{X} \left[ (\bar{\bm{h}}_n + \delta \bm{h}_n)^T (\bar{\bm{h}}_l + \delta \bm{h}_l) \right],
\end{align}
where $\bar{\bm{h}}_n=\mathbb{E}_{\bm{x}\in\mathcal{V}_t}[\bm{h}_n]$ denotes the local mean feature over the neighborhood $\mathcal{V}_t$, and $\delta\bm{h}_n=\bm{h}_n-\bar{\bm{h}}_n$ is the corresponding local deviation.

\par\smallskip
\noindent\textbf{Common Subspace Extraction and the Role of $U$.} To enable a geometry-aware comparison, we extract a \emph{common subspace} from the combined space of $\bm{h}_n$ and $\bm{h}_l$ by stacking $H = \begin{bmatrix} \bm{h}_n^T & \bm{h}_l^T \end{bmatrix}^T = U \Sigma V^T$. As detailed in Equation A.3, projecting onto the common subspace $U$ provides an orthogonal basis that summarizes correlated directions shared across layers within the local neighborhood. Based on observation in Figure \ref{fig:tsne}, we interpret the projected components as:
\begin{itemize}
\item $\delta \bm{h}_n^T \delta \bm{h}_n $ represents the residual component ($\bm{u}_{\text{residual}}$), dominated by layer-local (often domain-sensitive) variation;
\item $\delta \bm{h}_n^T \delta \bm{h}_l$ and $\delta \bm{h}_l^T \delta \bm{h}_n$ represent push-forward induced by transportation ($\bm{u}_{\text{push}}$), capturing cross-layer transport toward semantic features;
\item $\delta \bm{h}_n^\top \left(\Phi_{n}^{l}\right)^\top \Phi_{n}^{l} \delta \bm{h}_n$ provides metric of the push-forward transportation ($\bm{u}_{\text{metric}}$), reflecting local metric/curvature information.
\end{itemize}
While component-wise basis isolation is non-trivial in an online mixed stream, the shared subspace $U$ provides a batch-wise geometry that emphasizes correlated cross-layer directions and suppresses batch-specific noise. This geometry is used for similarity normalization in DFM.
\par\smallskip
\noindent\textbf{Domain-agnostic Flow Matching Loss.} With this insight, we guide intermediate representations away from domain-specific geometry and toward class-discriminative geometry.
Let $\bm{h}^*_n$ be a perturbed function of $\bm{h}_n$ with continuous deformation (e.g., induced by prompts); locally, we apply the same $U$ extracted above.
For a sample $\bm{x}_{(i)}$ in batch $\bm{X}$, let the intermediate feature $\bm{h}_{n,(i)} \in \bm{H}_n$, last-layer feature $\bm{h}_{l,(i)} \in \bm{H}_l$ from the frozen backbone model, and intermediate features from the perturbed function $\bm{h}^*_{n,(i)} \in \bm{H}_n^*$.
\par\smallskip
\noindent\textbf{Positive/Negative Design: Semantic Pull and Domain-Biased Push.}
We use the final-layer feature as a semantic anchor and set
$\bm{H}^{+}_{(i)} = \left\{ \bm{h}_{l,(i)} \right\}$ as the positive set.
To push $\bm{h}^*_{n,(i)}$ away from domain-sensitive intermediate regions (Fig.~\ref{fig:tsne-a}), we use frozen intermediate features as \emph{domain-biased negatives}.
Other samples' final-layer features are also included to prevent collapse and preserve instance-level discrimination. The negative set is
\begin{align}
\bm{H}^{-}_{(i)} =
\left\{ \bm{h}_{n,(j)} \mid \bm{h}_{n,(j)} \in \bm{H}_n \right\}
\cup
\left\{ \bm{h}_{l,(j)} \mid i \neq j \land \bm{h}_{l,(j)} \in \bm{H}_l \right\}.
\end{align}
\noindent We propose a contrastive loss formulated as follows:
\begin{align}
    & \mathcal{L}_{\text{DFM}} = -\frac{1}{M} \sum_{m=1}^{M} \log \frac
    {\sum_{\bm{h}_{n,(m)}^{+} \in \bm{H}^{+}_{(m)}} \exp\left(\text{cos}(\bm{h}^{*}_{(m)} U, \bm{h}_{(m)}^{+} U) / \tau\right)}
    {\sum_{\bm{h}_{n,(m)}^{+} \in \bm{H}^{+}_{(m)}} \exp\left(\text{cos}(\bm{h}^{*}_{(m)} U, \bm{h}_{(m)}^{-} U / \tau\right)},
    \label{eq:dfm}
\end{align}
where cos is cosine similarity and $\tau$ is a temperature, while dividing by norm reduces the effect of the amplitude of the feature. The further details of the derivation and procedure are provided in Appendix A.2 and Algorithm A.2.

\subsection{Global Topology Preservation}

\noindent One significant challenge in Online VIL is that the model must train from batches that only partially represent the overall distribution. This issue is particularly acute when input batches exhibit non-stationary class-domain compositions.
This challenge is particularly acute in the Online VIL scenario, where input batches exhibit non-stationary class-domain compositions.
To address this representational instability and to ensure topological coherence of the feature space, we introduce \textbf{Global Topology Preservation (GTP)}, a novel approach designed to maintain semantic structural integrity across temporally evolving data streams.
\par\smallskip
\noindent The effective information content of batch $B_t$ at layer $\bm{h}_l(\bm{x})$ can be measured by the rank of its empirical covariance matrix:
\begin{align}
    C_t = \mathbb{E}_{\bm{x} \in B_t} \left[ (\bm{h}_l(\bm{x}) - \mu_t)(\bm{h}_l(\bm{x}) - \mu_t)^T \right], \quad
    \mu_t = \mathbb{E}_{\bm{x} \in B_t} \left[ \bm{h}_l(\bm{x}) \right],
\end{align}
when certain classes are absent, $C_t$ captures only partial information about the global feature distribution, leading to rank deficiency.
When batch $B_t$ contains samples from only $k < C$ classes, the rank of the gradient is limited to $k$; the gradient provides not only the construction of decision boundaries for the present classes but also distorts the feature space by contracting around them.
This rank deficiency yields incomplete feature representations and biases gradients toward observed classes, causing the feature space to contract around present classes while neglecting absent ones.
This issue persists even with a prototype-based classifier \cite{snell2017prototypical}, which assumes each class is independent, thereby removing the absent rank into a zero-eigenvalue space.
\par\smallskip
\noindent To circumvent these limitations while preserving global topological properties, GTP constructs a surrogate representation of the feature manifold through two key components:
a set of $k$ global prototypes $\{\bar{p}_g^j\}_{j=1}^{k}$, and a set of global relationship vectors $\{\bar{r}_g^{j,l}\}_{j,l=1}^{k}$.
For each incoming batch $B_t$, we derive batch-specific prototypes $\{p_b^i\}_{i=1}^{k}$ from the final layer features using FINCH clustering \cite{finch}.
We employ an exponential moving average (EMA) update strategy to ensure smooth temporal evolution of global prototypes while mitigating batch-to-batch fluctuations.
Once the correspondence is established, each global prototype $\bar{p}_g$ is updated with the batch prototype $p_b$:
\begin{align}
    \bar{p}_g^{\pi^*(i), \text{new}} = (1 - \alpha) \bar{p}_g^{\pi^*(i), \text{old}} + \alpha p_b^i \quad \text{for } i = 1, \dots, k,
\end{align}
where $\alpha$ controls EMA update rate, balancing stability and adaptability.
We find optimal assignment $\pi^*$ by solving:
$\pi^* = \arg\min_{\pi \in S_k} \sum_{i=1}^{k} \|p_b^i - \bar{p}_g^{\pi(i)}\|$
using Hungarian algorithm \cite{kuhn1955hungarian}.
\par\smallskip
\noindent The relationships between its constituent elements fundamentally characterize the topological structure.
To capture these structural properties, We introduce a learnable mapping function $\phi:\mathbb{R}^{2d}\rightarrow\mathbb{R}^{m}$, where concatenation preserves both individual prototype and relative positioning. We define batch-specific relationship vectors $r_b^{i,j}=\phi([p_b^i;p_b^j])$ and global relationship vectors $\bar r_g^{i',j'}=\phi([\bar p_g^{i'};\bar p_g^{j'}])$. The GTP loss then aligns these relationship structures as follows:
\begin{equation}
    \mathcal{L}_{\text{GTP}} = \sum_{i=1}^{k} \sum_{\substack{j=1 \\ j \neq i}}^{k} D (r_b^{i,j}, \bar{r}_g^{\pi^*(i),\pi^*(j)}), \label{eq:gtp_loss}
\end{equation}
where $D$ represents the cosine distance metric, this formulation encourages consistent pairwise relationships between semantic prototypes, effectively preserving the global topological structure without explicitly computing spectral properties.
\par\smallskip
\noindent Since the number of global prototypes $\bar{p}_g$ is smaller than the number of samples in each batch, these prototypes deliberately capture a coarse-grained representation, reflecting the intended vagueness under limited observations.
Each prototype is updated considering its relationships with all other prototypes.
These prototypes cover the feature space of several semantic classes, which reduces the distortion of the feature space from the concept without samples.
Additionally, the EMA update strategy stabilizes the prototypes over time and removes outdated information from previous batches.
As shown in Figure A.3 and A.4 in Appendix, this effectively filters out the unseen class information in recent batches, performing a role similar to a short-term memory.
This mechanism provides a computationally efficient solution to maintaining representational stability in non-stationary environments, complementing the domain-invariance properties induced by DFM.

%% file: sec/4_experiment.tex
\section{Experiments}
\label{experiment}
\noindent This section evaluates and compares proposed TopFlow against state-of-the-art methods.
Section \ref{sec:experimental setup} describes the experimental setup, including datasets, baselines, and evaluation metrics. Section \ref{sec:experimental results} presents extensive quantitative results, demonstrating the effectiveness of TopFlow across multiple benchmarks. To further analyze TopFlow, Section \ref{sec:ablation} provides in-depth analyses, including ablation studies, to isolate and assess the contribution of each component in TopFlow. Further details about the experiments are provided in the Appendix.

\subsection{Experimental Setup}
\label{sec:experimental setup}
\noindent\textbf{Datasets.}
We conducted experiments on three benchmarks, including iDigits \cite{volpi2021continual}, CORe50 \cite{lomonaco2017core50}, and CLEAR100 \cite{lin2021clear}, for which it is possible to construct Online VIL scenarios that can cause a significant shift in distribution by clearly distinguishing both classes and domains. We split the CORe50 and CLEAR100 datasets into 10 tasks for the task construction, and 5 for the iDigits dataset.
\par\smallskip
\noindent\textbf{Baselines.}
We compared our proposed method with traditional naive baselines and the latest state-of-the-art methods.
First, we set the lower bound as the usual supervised sequential fine-tuning result (FT) and the upper bound as the usual supervised joint fine-tuning result.
Then, we compared our proposed method with replay-based methods such as ER \cite{rolnick2019experience}, Rainbow Memory (RM) \cite{bang2021rainbow}, CLIB \cite{koh2021online}, and CBA \cite{wang2023cba}, regularization-based method EWC \cite{kirkpatrick2017overcoming}, LwF \cite{li2017learning}, SLCA \cite{zhang2023slca}, DYSON \cite{he2024dyson}, OCM \cite{guo2022online}, OnPro \cite{wei2023online}, PEC \cite{zajkac2024prediction}, S6MOD \cite{liu2025enhancing}, DUCT \cite{zhou2025dual} and prompt-based CODA-P \cite{smith2023coda}, ICON \cite{park2024versatile} and MVP \cite{moon2023online}.
\par\smallskip
\noindent\textbf{Implementation Details.}
We used the MVP \cite{moon2023online} as our baseline for model and experimental setups.
We used Adam optimizer with a learning rate of 5e-3, and implemented with a batch size of 64.
As a mapping function $\psi$ which maps prototypes into a relation vector, we used a simple 2-layer MLP function.
The hidden dimension of $\psi$ is 64 in CORe50, and 32 in CLEAR100.
The size of the dimension of the relation vector $m$ is 10.
For the EMA update of global feature topology, a decay factor of 0.99 was used.
Experiments were conducted under assumption of the memory-free setting, we adopted naïve reservoir memory for the experiments with memory buffer. We conducted experiments with 3 random seeds, and note that using more seeds (e.g., 10 runs) also yields consistent results. Details are described in the Appendix.
\par\smallskip
\noindent\textbf{Evaluation Metrics.}
To evaluate online learning performance, we employed two metrics: $A_\text{AUC}$ and $A_\text{Last}$ \cite{koh2021online}.
The $A_\text{AUC}$ metric quantifies performance under anytime inference, where inference queries may occur at arbitrary points during training as new classes are encountered.
Conversely, $A_\text{Last}$ assesses inference accuracy after training.
In real-world applications, models must deliver reliable predictions on demand, regardless of training stage.
Thus, $A_\text{AUC}$ and $A_\text{Last}$ provide a robust framework for benchmarking online learning performance.

\subsection{Experimental Results}
\label{sec:experimental results}
\input{tab/main}
\noindent We conducted extensive experiments in the proposed Online VIL scenario, and the results are summarized in Table \ref{tab:main} and Table \ref{tab:main_memory_full}.
As shown in Table 1, under the non-replay setting, TopFlow consistently outperforms existing methods in both $A_\text{AUC}$ and $A_\text{Last}$.
These results demonstrate that our approach mitigates catastrophic forgetting while enabling continual adaptation to the input stream.  Moreover, in Table \ref{tab:main_memory_full}, the introduction of replay memory generally improves performance, but the proposed TopFlow consistently outperforms all baselines.
Surprisingly, in replay-buffer settings, we observe that naïve methods such as Experience Replay (ER) and earlier methods tend to achieve the best performance, except for our proposed method.
This suggests that most existing online-incremental learning methods struggle in realistic Online VIL scenarios where distribution shifts are frequent and task boundaries are unclear.
In contrast, our proposed TopFlow maintains strong performance across all settings, confirming its robustness. Furthermore, TopFlow achieves a more stable accuracy trajectory over time, with fewer drastic performance drops between tasks. This indicates that DFM and GTP help smooth the learning process by leveraging more structured representations and effective feature alignment.

\subsection{Ablation Studies and Analysis}

\input{tab/main_memory}
\input{tab/ablation}

\label{sec:ablation}

\noindent\textbf{Ablation Study.}
Table \ref{tab:ablation} demonstrates the effectiveness of each component in our TopFlow framework.
Implementing DFM or GTP individually significantly enhanced both $A_\text{AUC}$ and $A_\text{Last}$, validating their respective contributions.
DFM stabilizes learning by distilling domain-invariant self-knowledge and enhancing class discrimination in later layers.
GTP preserves performance under varying batch compositions by consolidating transient batch cues into a persistent feature topology.
Together, they provide complementary gains: DFM improves representations, while GTP maintains their structure for Online VIL.
\par\smallskip
\noindent\textbf{Effectiveness of TopFlow in Standard CL Scenarios.}
As shown in Table \ref{tab:accuracy_variants}, DFM and GTP improve performance in both standard CIL and DIL, individually and combined.
We adopt CIFAR-100 \cite{krizhevsky2009learning} 10-split CIL and CORe50 \cite{lomonaco2017core50} 8-split DIL as standard setups, with CODA-P \cite{smith2023coda} and S-Prompts \cite{wang2022sprompts} as baselines, respectively.
Despite the limited domain variation in CIL, both components provide consistent gains, and the improvements are more pronounced in DIL, supporting their motivation under domain shifts.

\input{tab/model_variants}
\par\smallskip
\noindent\textbf{Layer Selection for DFM.}
To analyze the effect of layer selection for Eq.~\ref{eq:dfm}, we conduct an ablation study using various $(n,l)$ pairs in the ViT encoder (Table \ref{tab:layer_selection}).
Applying DFM to early layers such as (0,5) degrades performance, suggesting that low-level features are not well-aligned with the high-level semantics captured in the final layer.
In contrast, using deeper layers such as (5,10) or (6,11) improves performance; combining (5,10) and (6,11) reaches 53.49, and our final configuration (6,11) achieves the best $A_\text{Last}$ of 54.82.
This indicates that later layers better preserve semantic information suitable for matching with final representations, yielding a favorable trade-off between abstraction and compatibility.
\par\smallskip
\noindent\textbf{DFM and GTP with Other Models.}
Since DFM and GTP are designed as general components, we further examine whether they remain effective when integrated with other CL algorithms. We conducted ablation experiments to verify whether the proposed DFM and GTP could also yield performance improvements on baseline CL algorithms other than the MVP. As shown in Table \ref{tab:method_variants}, DFM and GTP lead to performance improvements over the baseline, even independently. DFM and GTP robustly improve performance over the baseline on the CORe50 dataset and other existing continual learning algorithms used in the main experiments.

\input{tab/offline_vil}
\par\smallskip
\noindent\textbf{Effectiveness of TopFlow in Offline VIL Scenarios.} As shown in Table \ref{tab:offline_vil}, the proposed TopFlow is effective not only in standard Offline CL scenarios such as CIL and DIL, but also in the more challenging Offline VIL setting, where both class and domain distributions change simultaneously. TopFlow achieves this by explicitly maintaining structural consistency by enforcing topology-preserving and Domain-agnostic matching across tasks, allowing newly learned representations to align with the previously learned feature structure. As a result, TopFlow achieves stable representation learning and consistently improves performance across Offline VIL benchmarks. From a scenario perspective, TopFlow was originally designed for the more challenging \textit{Online VIL} setting, where the model must adapt to non-stationary data streams without access to the full dataset distribution. Nevertheless, as shown in Table \ref{tab:offline_vil}, TopFlow also achieves strong performance in the \textit{Offline VIL} scenario. In contrast, the Offline VIL method ICON fails to maintain competitive performance in the Online VIL setting, as reported in Table \ref{tab:main_memory_full}. This result suggests that while methods tailored to Offline VIL may not generalize well to streaming environments, TopFlow provides a more robust solution that transfers across different VIL scenarios.

\begin{figure}[!t]
    \centering
    \includegraphics[width=0.95\linewidth]{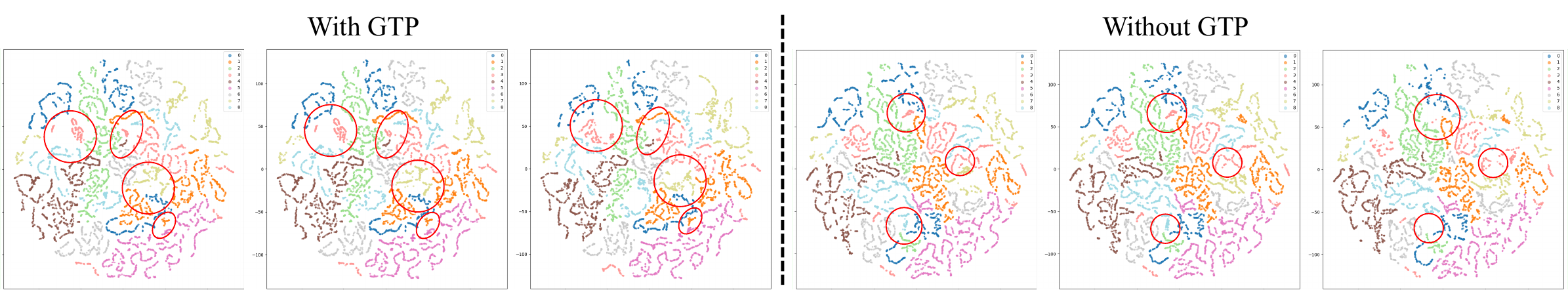}
    \caption{t-SNE visualization of output features over tasks with GTP and without GTP.}
    \label{fig:gtp_tsne} 
\end{figure}
\par\smallskip
\noindent\textbf{Visualization of GTP.} To qualitatively analyze the effect of GTP, we visualize the output features using t-SNE \cite{JMLR:v9:vandermaaten08a} over tasks with and without GTP in Figure \ref{fig:gtp_tsne}. We tracked the evolution of features from the first 10 classes, which are available for all tasks in the CORe50 dataset.
And low-dimensional approximation is performed across all tasks to maintain consistency and relative positions of features. While it is challenging to project high-dimensional topological structures into 2D space, we observe that GTP helps to preserve the relative arrangement and internal structure of clusters across tasks. Especially in red circles, the meaningful topologies (e.g., relative connections or penetrations between clusters) are better maintained with GTP, whereas they drift and dismorph without it. This qualitative analysis supports our findings, demonstrating that GTP preserves global topological relationships in the feature space during online learning.

%% file: tab/main.tex
\begin{table}[!t]
\caption{Experimental results with proposed Online VIL scenarios. We used bold and underlined as brief indications of the best and the second best, respectively.}
\centering
\resizebox{\linewidth}{!}{
    \centering
    \begin{tabular}{cccccccccc}
    \specialrule{1.2pt}{0pt}{1.2pt}
       \multirow{2}{*}{\fontseries{b}\selectfont{Method}} &  & \multicolumn{2}{c}{\fontseries{b}\selectfont{iDigits}} &  & \multicolumn{2}{c}{\fontseries{b}\selectfont{CORe50}} &  & \multicolumn{2}{c}{\fontseries{b}\selectfont{CLEAR100}} \\ \cline{3-4}\cline{6-7}\cline{9-10}
         &  & $A_\text{AUC}$ & $A_\text{Last}$ &  & $A_\text{AUC}$ & $A_\text{Last}$ &  & $A_\text{AUC}$ & $A_\text{Last}$  \\ \hline
Upper-bound & & - & 87.18$\pm$0.13 & & - & 91.66$\pm$0.25 & & - & 94.36$\pm$0.28 \\
Lower-bound & & 13.45$\pm$0.62 & 12.71$\pm$3.34 & & 3.39$\pm$0.10 & 3.24$\pm$1.09 & & 2.38$\pm$0.32 & 2.66$\pm$0.55 \\ \cdashline{1-10}
EWC \cite{kirkpatrick2017overcoming} & & 20.07$\pm$2.84 & 14.67$\pm$3.61 & & 18.60$\pm$4.64 & 16.07$\pm$1.56 & & 23.61$\pm$3.11 & 19.93$\pm$1.33 \\
LwF \cite{li2017learning} & & 19.61$\pm$3.50 & 15.38$\pm$1.04 & & 25.27$\pm$3.77 & 21.97$\pm$4.22 & & 23.80$\pm$2.24 & 21.70$\pm$4.19 \\
CODA-P \cite{smith2023coda}& & 23.96$\pm$4.74 & 20.62$\pm$3.47 & & 54.06$\pm$5.32 & 48.88$\pm$2.92 & & 28.82$\pm$5.77 & 25.61$\pm$3.52  \\
SLCA \cite{zhang2023slca}& & 35.81$\pm$3.98 & 24.88$\pm$2.82 & & 33.49$\pm$4.47 & 27.36$\pm$2.06 & & 32.16$\pm$2.28 & 31.70$\pm$1.13 \\
PEC \cite{zajkac2024prediction}& & 34.77$\pm$3.02 & 28.01$\pm$2.69 & & 51.35$\pm$4.39 & 46.95$\pm$2.28 & & 53.93$\pm$3.20 & 51.66$\pm$2.09 \\
ICON \cite{park2024versatile} & & 33.60$\pm$2.16 & 30.63$\pm$2.77 & & 49.42$\pm$3.29 & 45.15$\pm$2.94 & & 59.38$\pm$2.46 & 58.60$\pm$2.57 \\
S6MOD \cite{liu2025enhancing} & & 31.18$\pm$1.62 & 30.42$\pm$2.55 & & 52.33$\pm$2.74 & 47.20$\pm$3.15 & & 59.60$\pm$2.79 & 57.47$\pm$2.48 \\
DUCT \cite{zhou2025dual} & & 36.46$\pm$3.48 & 30.94$\pm$2.41 & & 53.66$\pm$4.37 & 47.24$\pm$1.28 & & 66.79$\pm$2.94 & 65.46$\pm$3.40 \\
MVP \cite{moon2023online}& & \underline{38.29$\pm$5.74} & \underline{31.05$\pm$3.15} & & \underline{58.30$\pm$4.48} & \underline{52.84$\pm$1.17} & & \underline{79.73$\pm$3.59} & \underline{77.11$\pm$2.33} \\ \cdashline{1-10}
\rowcolor{lightgray} \fontseries{b}\selectfont{TopFlow (Ours)}& &\fontseries{b}\selectfont{48.52}$\pm$\fontseries{b}\selectfont{1.25} &\fontseries{b}\selectfont{32.18}$\pm$\fontseries{b}\selectfont{1.01} & &\fontseries{b}\selectfont{64.51}$\pm$\fontseries{b}\selectfont{2.50} &\fontseries{b}\selectfont{66.20}$\pm$\fontseries{b}\selectfont{4.18}&  &\fontseries{b}\selectfont{87.12}$\pm$\fontseries{b}\selectfont{0.01}  &\fontseries{b}\selectfont{80.64}$\pm$\fontseries{b}\selectfont{2.67}  \\
\specialrule{1.2pt}{0pt}{0pt}
\end{tabular}
}
\label{tab:main}
\end{table}

%% file: tab/main_memory.tex
\begin{table*}[!t]
\caption{Results of OnlineVIL scenarios using replay buffer sizes 500 and 2000.}
\centering
\resizebox{\linewidth}{!}{
    \centering
    \begin{tabular}{cccccccccccc}
    \specialrule{1.1pt}{1pt}{1pt}
       \multirow{2}{*}{\begin{tabular}{c}\fontseries{b}\selectfont{Buffer} \\ \fontseries{b}\selectfont{Size}\end{tabular}}& \multirow{2}{*}{\fontseries{b}\selectfont{Method}} &  & \multicolumn{2}{c}{\fontseries{b}\selectfont{iDigits}} &  & \multicolumn{2}{c}{\fontseries{b}\selectfont{CORe50}} &  & \multicolumn{2}{c}{\fontseries{b}\selectfont{CLEAR100}} & \\ \cline{4-5}\cline{7-8}\cline{10-11}
         &  &  & $A_\text{AUC}$ & $A_\text{Last}$ &  & $A_\text{AUC}$ & $A_\text{Last}$ &  & $A_\text{AUC}$ & $A_\text{Last}$ &  \\ \hline
\multirow{9}{*}{500} & ER \cite{rolnick2019experience} & & \underline{59.43$\pm$6.24} & 48.70$\pm$2.51 & & 74.77$\pm$4.85 & 72.25$\pm$2.27 & & 73.92$\pm$3.93 & 71.49$\pm$3.20 & \\
         & RM \cite{bang2021rainbow} & & 55.02$\pm$5.37 & 51.73$\pm$2.05 & & 81.06$\pm$3.90 & 70.41$\pm$3.17 & & 72.42$\pm$4.64 & 72.93$\pm$2.05 & \\
         & CLIB \cite{koh2021online} & & 57.38$\pm$4.16 & 52.63$\pm$3.38 & & 75.06$\pm$5.81 &  71.93$\pm$1.06 & & 68.39$\pm$5.25 & 66.92$\pm$1.52 & \\
         & OCM \cite{guo2022online}& & 57.40$\pm$3.60 & 52.88$\pm$2.52 & &75.29$\pm$3.10 & 72.66$\pm$1.93 & & 77.80$\pm$3.25 & 75.10$\pm$2.42 & \\
         & CBA \cite{wang2023cba}& & 58.05$\pm$4.39 & \underline{54.28$\pm$3.07} & & 81.92$\pm$4.04 & \underline{81.02$\pm$1.39} & & 75.26$\pm$4.08 & 74.47$\pm$1.82 & \\
         & OnPro \cite{wei2023online} & & 46.92$\pm$4.82 & 44.86$\pm$1.63 & & 71.39$\pm$4.03 & 70.92$\pm$2.24  & & 81.36$\pm$4.99 & 77.46$\pm$1.53 & \\
         & DYSON \cite{he2024dyson} &  & 42.31$\pm$3.11 & 38.18$\pm$3.56 & & 62.92$\pm$5.61 & 60.72$\pm$2.16 & & 66.56$\pm$4.65 & 65.62$\pm$2.74 & \\
         & MVP-R \cite{moon2023online} &  & 48.29$\pm$3.73 & 40.97$\pm$2.15 & & \underline{83.26$\pm$5.16} & 80.16$\pm$1.03 & & \underline{87.82$\pm$3.17} & \underline{85.65$\pm$2.18} & \\ \cdashline{2-11}
\rowcolor{lightgray}\cellcolor{white} & \fontseries{b}\selectfont{TopFlow (Ours)} && \fontseries{b}\selectfont{62.48}$\pm$\fontseries{b}\selectfont{4.17} & \fontseries{b}\selectfont{56.48}$\pm$\fontseries{b}\selectfont{1.52}  &  & \fontseries{b}\selectfont{85.16}$\pm$\fontseries{b}\selectfont{0.84}  & \fontseries{b}\selectfont{91.14}$\pm$\fontseries{b}\selectfont{0.05} &  & \fontseries{b}\selectfont{91.67}$\pm$\fontseries{b}\selectfont{0.03} & \fontseries{b}\selectfont{90.12}$\pm$\fontseries{b}\selectfont{1.00} & \\ \hline
\multirow{9}{*}{2000} & ER \cite{rolnick2019experience}&  & \underline{61.72$\pm$5.12} & \underline{57.81$\pm$1.47} &  & 79.44$\pm$5.17 & 76.61$\pm$3.91 &  & 83.51$\pm$4.61 & 81.03$\pm$3.59 & \\
         & RM \cite{bang2021rainbow} &  & 43.96$\pm$6.18 & 41.23$\pm$3.62 &  &82.42$\pm$3.03 & 79.84$\pm$2.19 & & 84.73$\pm$4.63 & 81.49$\pm$2.26 & \\
         & CLIB \cite{koh2021online} &  & 51.96$\pm$3.81 & 46.06$\pm$2.34 &  &84.58$\pm$4.26 & 81.63$\pm$2.51 & &  85.62$\pm$5.05 & 83.25$\pm$1.62 \\
         & OCM \cite{guo2022online}&  & 57.52$\pm$5.08 & 50.60$\pm$3.34 &  &84.92$\pm$4.03 & 83.24$\pm$2.72 & &  84.26$\pm$4.82 & 82.91$\pm$3.56 \\
         & CBA \cite{wang2023cba}&  & 60.02$\pm$4.52 & 55.04$\pm$2.82 &  &85.16$\pm$3.28 & \underline{83.49$\pm$3.20} & &  88.02$\pm$3.80 & 85.94$\pm$2.36   \\
         & OnPro \cite{wei2023online} &  & 54.82$\pm$4.95 & 51.03$\pm$3.77 &  & 81.35$\pm$5.51 & 78.09$\pm$3.91 & &  88.83$\pm$5.16 & 86.11$\pm$3.57   \\
         & DYSON \cite{he2024dyson} &  & 46.74$\pm$4.41 & 43.38$\pm$4.28 &  & 51.21$\pm$3.71 & 49.29$\pm$1.74 &  & 57.05$\pm$4.18 & 55.48$\pm$3.43 & \\
         & MVP-R \cite{moon2023online} &  & 52.14$\pm$2.99 & 47.74$\pm$2.36 & & \underline{87.33$\pm$3.37} & 82.39$\pm$1.10 & & \underline{89.48$\pm$1.71} & \underline{88.93$\pm$1.18} & \\ \cdashline{2-11}
\rowcolor{lightgray}\cellcolor{white} & \fontseries{b}\selectfont{TopFlow (Ours)} && \fontseries{b}\selectfont{65.45}$\pm$\fontseries{b}\selectfont{3.84} & \fontseries{b}\selectfont{60.39}$\pm$\fontseries{b}\selectfont{4.28} &  & \fontseries{b}\selectfont{87.56}$\pm$\fontseries{b}\selectfont{0.82} &\fontseries{b}\selectfont{92.24}$\pm$\fontseries{b}\selectfont{0.15}  &  &\fontseries{b}\selectfont{93.55}$\pm$\fontseries{b}\selectfont{0.02}& \fontseries{b}\selectfont{92.97}$\pm$\fontseries{b}\selectfont{0.65}  & \\ 
    \specialrule{1.1pt}{1pt}{1pt}
    \end{tabular}
}
    \label{tab:main_memory_full}
\end{table*}

%% file: tab/ablation.tex
\begin{table}[t]
\centering
\begin{minipage}[t]{.29\linewidth}
    \centering
    \caption{Ablation study for DFM and GTP on CORe50.}
    \vspace{-0.2cm}
    \resizebox{\linewidth}{!}{
        \begin{tabular}[t]{cccc}
        \specialrule{1.2pt}{0pt}{1.2pt} 
        DFM & GTP & $A_\text{AUC}$ & $A_\text{Last}$ \\
        \hline
        \multicolumn{2}{c}{Baseline} &  58.30 & 52.84  \\
        \cdashline{1-4}
        $\checkmark$ &  & 64.13  & 64.57 \\
        & $\checkmark$  & 63.95 & 65.28 \\ \cdashline{1-4}
        $\checkmark$ & $\checkmark$ & \textbf{64.51} & \textbf{66.20} \\
        \specialrule{1.2pt}{0pt}{1.2pt} 
        \end{tabular}
    }
    \label{tab:ablation}
\end{minipage}
$\;$
\begin{minipage}[t]{.39\linewidth}
    \centering
    \caption{The Effectiveness of TopFlow in Standard CL Scenarios.}
    \vspace{-0.2cm}
    \resizebox{\linewidth}{!}{
        \begin{tabular}[t]{ccc}
        \specialrule{1.2pt}{0pt}{1.2pt} 
        \multirow{2}{*}{\textbf{Method}} & \multicolumn{2}{c}{\textbf{Accuracy}} \\
        & \textbf{CIL, CODA-P} & \textbf{DIL, S-Prompt} \\
        \hline
        Baseline & 84.17 & 82.96\\
        \cdashline{0-2}
        + DFM & 84.19 & 85.36 \\
        + GTP & 85.52 & 86.88 \\
        \cdashline{0-2}
        \textbf{+ DFM, GTP} & \textbf{86.04} & \textbf{87.50} \\
        \specialrule{1.2pt}{0pt}{1.2pt} 
        \end{tabular}
    }
    \label{tab:accuracy_variants}
\end{minipage}
$\;$
\begin{minipage}[t]{.25\linewidth}
    \centering
    \caption{Ablation of layer selection in the DFM loss (w/o GTP). }
    \vspace{-0.2cm}
    \resizebox{\linewidth}{!}{
    \begin{tabular}[t]{cc}
    \specialrule{1.2pt}{0pt}{1.2pt} 
    Layer   & $A_\text{Last}$    \\ \hline
    Baseline & 52.84 \\ \cdashline{1-2}
    (0, 5) & 49.92 \\
    (5, 10) & 52.98 \\
    (5, 10), (6, 11) & 53.49 \\
    \textbf{(6, 11) (Ours)}& \textbf{54.82} \\
    \specialrule{1.2pt}{0pt}{1.2pt} 
    \end{tabular}
    }
    \label{tab:layer_selection}
\end{minipage}
\vspace{-0.2cm}
\end{table}

%% file: tab/model_variants.tex
\setlength\intextsep{-3pt}
\parskip=0pt
\begin{wraptable}{R}{0.25\textwidth}
\centering
\caption{Performance comparison of different methods and variants.}
\resizebox{\linewidth}{!}{
\begin{tabular}{cc}
\specialrule{1.2pt}{0pt}{1.2pt} 
Method& \textbf{$A_\text{Last}$} \\
\hline
CODA-P & 49.13 \\
+ DFM & 52.88 \\
+ GTP & 53.17 \\
\textbf{+ DFM, GTP} & \textbf{54.62} \\
\hline
PEC & 45.99 \\
+ DFM & 46.64 \\
+ GTP & 48.19 \\
\textbf{+ DFM, GTP} & \textbf{48.86} \\
\specialrule{1.2pt}{0pt}{1.2pt} 
\end{tabular}
}
\label{tab:method_variants}
\end{wraptable} 

%% file: tab/offline_vil.tex
\begin{table}[!t]
\caption{The Effectiveness of TopFlow in Offline VIL Scenarios.}
\vspace{-0.2cm}
\centering
\resizebox{\linewidth}{!}{
    \centering
    \begin{tabular}{cccccccccc}
    \specialrule{1.2pt}{0pt}{1.2pt}
       \multirow{2}{*}{\textbf{Method}} &  & \multicolumn{2}{c}{\textbf{iDigits}} &  & \multicolumn{2}{c}{\textbf{CORe50}} &  & \multicolumn{2}{c}{\textbf{DomainNet}} \\ \cline{3-4}\cline{6-7}\cline{9-10}
         &  & Avg. Acc $\uparrow$  & Forgetting $\downarrow$  &  & Avg. Acc $\uparrow$  & Forgetting $\downarrow$ &  & Avg. Acc $\uparrow$  & Forgetting $\downarrow$  \\ \hline
ICON & & 70.67$\pm$2.13 & 13.49$\pm$2.09 && 79.61$\pm$1.73 & 8.19$\pm$1.47 && 49.08$\pm$2.11 & 17.88$\pm$2.40 \\
\fontseries{b}\selectfont{TopFlow (Ours)} & & \fontseries{b}\selectfont{73.33$\pm$1.72} & \fontseries{b}\selectfont{12.14$\pm$4.20} && \fontseries{b}\selectfont{79.77$\pm$1.47} & \fontseries{b}\selectfont{8.03$\pm$1.19} && \fontseries{b}\selectfont{49.62$\pm$1.82} & \fontseries{b}\selectfont{17.30$\pm$3.09} \\
\specialrule{1.2pt}{0pt}{0pt}
\end{tabular}
}
\label{tab:offline_vil}
\end{table}

%% file: sec/5_conclusion.tex
\section{Conclusion}
\noindent We proposed a new online continual learning scenario named Online VIL, which simulates a complex real world where states are ever-changing and there are no concepts of tasks and clear boundaries between them.
Through analysis, we determined the direction for problem-solving in Online VIL and defined novel TopFlow framework.
We demonstrated that the proposed TopFlow showed SOTA performance in the challenging Online VIL scenario, and its effectiveness through various experiments.
Online VIL involves stochastic task construction, where the composition of data to each task is influenced by random seed. While this design captures more realistic dynamics, it can introduce variability in results. Despite this, we hope that our Online VIL scenario will serve as a new benchmark for advancing real-world incremental learning research, providing a more realistic and challenging setting for future studies. 